\documentclass[letterpaper, journal, 10pt]{IEEEtran}
\usepackage{cite}
\usepackage{amsmath,amssymb,amsfonts}
\usepackage{algorithmic}
\usepackage{graphicx}
\usepackage{textcomp}
\usepackage{xcolor}
\usepackage{soul}
\usepackage{comment}
\usepackage{bm}
\usepackage{multirow}
\usepackage{dsfont}
\usepackage{ulem}
\usepackage[tight]{subfigure}

\newtheorem{Def}{Definition}

\begin{document}

\title{Toward Robotic Weed Control: Detection of Nutsedge Weed in Bermudagrass Turf Using Inaccurate and Insufficient Training Data
\author{Shuangyu Xie, Chengsong Hu, Muthukumar Bagavathiannan, and Dezhen~Song     
\thanks{S. Xie and C. Hu are co-first authors of this paper. S. Xie is supervised by D. Song at the Computer Sci. and Eng. Dept.; C. Hu is supervised by  M. Bagavathiannan at the Dept. of Soil and Crop Sciences,  Texas A\&M University, College Station, TX 77843 US. 
	(Emails: \textit{sy.xie@tamu.edu; huchengsong@tamu.edu; muthu@tamu.edu; and dzsong@cs.tamu.edu}). 
}
\thanks{This work was supported in part by TAMU and NSF under NRI-1925037.}
}

}

\maketitle

\begin{abstract}
To enable robotic weed control, we develop algorithms to detect nutsedge weed from bermudagrass turf. Due to the similarity between the weed and the background turf, manual data labeling is expensive and error-prone. Consequently, directly applying deep learning methods for object detection cannot generate satisfactory results. Building on an instance detection approach (i.e. Mask R-CNN), we combine synthetic data with raw data to train the network. We propose an algorithm to generate high fidelity synthetic data, adopting different levels of annotations to reduce labeling cost. Moreover, we construct a nutsedge skeleton-based probabilistic map (NSPM) as the neural network input to reduce the reliance on pixel-wise precise labeling. We also modify loss function from cross entropy to Kullback–Leibler divergence which accommodates uncertainty in the labeling process. We implement the proposed algorithm and compare it with both Faster R-CNN and Mask R-CNN. The results show that our design can effectively overcome the impact of imprecise and insufficient training sample issues and significantly outperform the Faster R-CNN counterpart with a false negative rate of only 0.4\%. In particular, our approach also reduces labeling time by 95\% while achieving better performance if comparing with the original Mask R-CNN approach.
\end{abstract}


\section{Introduction}

We are interested in developing robotic weed removal solutions for environmentally-friendly lawn care. One key issue is to be able to recognize weeds from background turfgrass using a low-cost camera on-board a robot. In this paper, we start with a particular instance: detection of nutsedge weed ({\em Cyperus spp}.; mix of yellow and purple nutsedges) in bermudagrass ({\em Cynodon dactylon}) turf.

However, weed detection is nontrivial. To an untrained eye, distinguishing a nutsedge plant from a turfgrass background is difficult especially in a recently mown lawn. Hence the manual data labeling process is expensive and error-prone. The resulting imprecise and insufficient training data is expected to significantly reduce the performance of common data-driven deep learning approaches.  

\begin{figure}[htb!]
 \centering
 \includegraphics[width=3.5in]{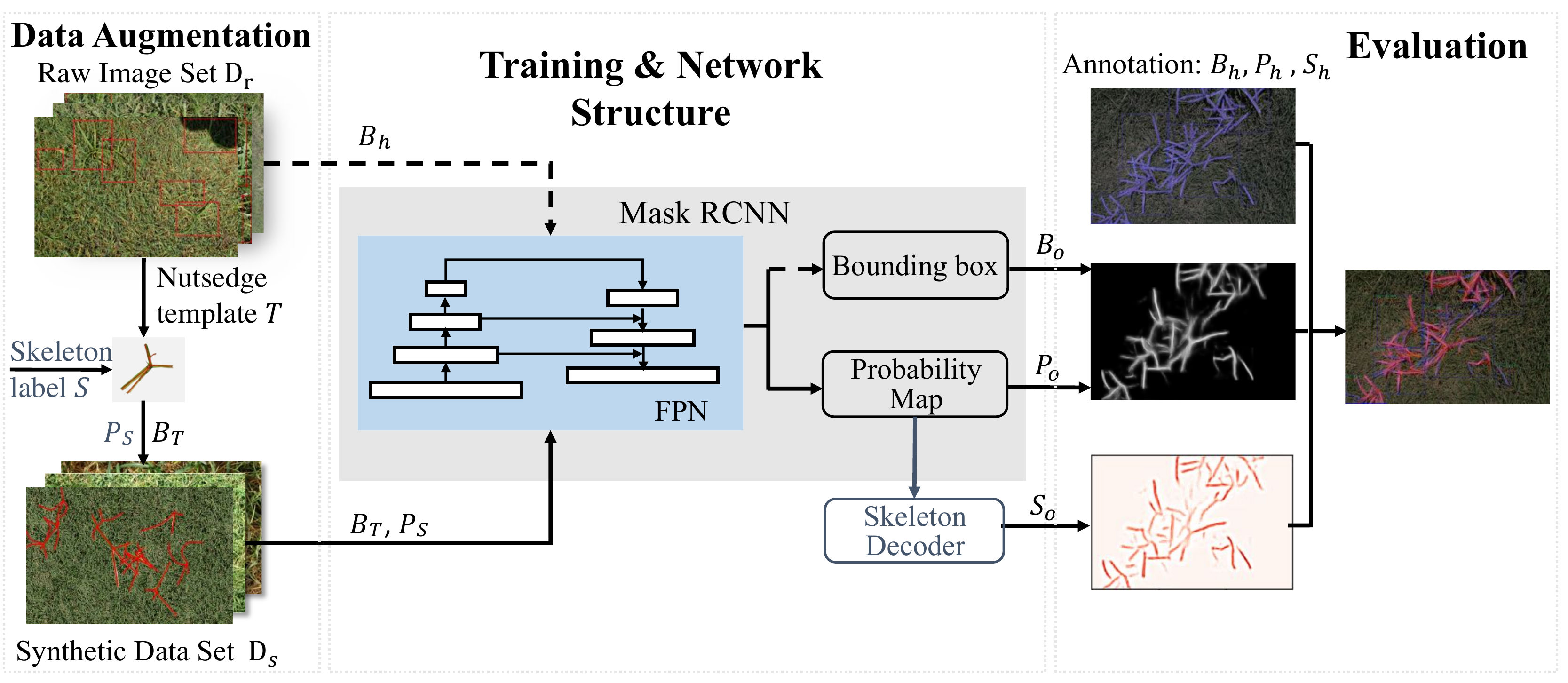}
\caption{An overview of nutsedge detection algorithms.}\vspace*{-.2in}
\label{fg:pipeline}
\end{figure}

Fig.~\ref{fg:pipeline} illustrates how we handle the challenge. First, we propose a data augmentation approach to allow us to combine synthetic data with raw data for neural network training. This significantly reduces the labeling requirement. We propose a data synthesis algorithm to generate high fidelity synthetic data, which also provides accurate labeling. Second, instead of relying on precise pixel-wise labeling, we employ annotations at different levels including bounding box and skeleton model to reduce labeling rigor requirement. Additionally, we propose a nutsedge skeleton-based probabilistic map (NSPM) representation. NSPM (e.g. $P_S$ in Fig.~\ref{fg:pipeline}) gives more weightage to the structure of nutsedge instead of equal treatment of individual pixels. Third, we modify our neural network loss function from cross entropy, which assumes accurate training samples, to Kullback–Leibler (KL) divergence, which measures the similarity between two probability functions that can take uncertainty in labeling into consideration. At last, we also propose new evaluation metrics to handle imprecise human labeling by extending existing intersection over union (IoU) metric and proposing a new skeleton similarity metrics using NSPM. We incorporate these new designs in a Mask R-CNN framework~\cite{mask} to complete our detection algorithm.

We have implemented the proposed algorithm and compared it with state-of-the-art methods such as Faster R-CNN~\cite{RCNN} and Mask R-CNN. The experimental results have shown that our algorithm significantly outperforms the counterparts. More specifically, the combination of using synthetic data with fine grain labels and raw image data with noisy bounding box labels under KL-divergence loss function leads to the lowest false negative rate of 0.4\%. In particular, our approach also reduces labeling time by 95\% while achieving better performance if comparing with the original Mask R-CNN approach.        


\section{Related Work}

Our work relates to robotic weed control, weed detection, image-based detection and segmentation, and data synthesis. 

\emph{Robotic weed control:} Recently, autonomous robots have seen many applications in precision agriculture because they have enormous potential to reduce operating costs and dependency on labor \cite{Multirobot,Pruning}. A robotic weed control system often includes three components: a sensing system to detect weeds, a decision-making unit to process the information from the sensing system and make manipulation decisions, and actuators to act accordingly \cite{Slaughter_robotic_weed_control}. Our work belongs to the first component~\cite{abidine_gps}. For robot decision making, it is important to localize individual plants, target the weeds and avoid crop plants ~\cite{english_row_detection}.  For actuation, selection of the actuation (weed-killing) mechanism is under fast development. Common actuation methods include cultivation tools\cite{mccool_robotic_weeding_tools}, stamping \cite{michaels_stamping}, mowing \cite{melita_mow}, precise herbicide application \cite{lee_spray} \cite{midtiby_spray}, etc. In addition to the actuator development, modular robotic platforms that are able to carry various weeding actuators are also under active development \cite{bawden_robot_platform}.

\emph{Image-based weed detection and segmentation:} In this area, methods can be categorized into two types: traditional computer vision methods and learning-based methods. Our weed detection algorithm developed here belongs to the latter. Before learning-based methods are widely adopted in solving weed detection problems, traditional computer vision methods that extract hand-craft plant visual characteristics have been commonly used. These characteristics can be classified into two major groups: visual texture and biological morphology \cite{weed_detection_review}. For example, Burks et al. \cite{burks_weed_texture} utilize the color co-occurrence method to discriminate textures between five common weed species. Herrera et al. \cite{herrera_weed_shape} propose a strategy utilizing a set of shape descriptors to discriminate grasses from broad-leaf weeds, which works when weeds are at an early stage of growth. 



Convolution neural network (CNN)-based methods outcompete traditional computer vision-based methods in feature extraction and have become more popular for weed detection nowadays. Many previous works employ CNNs to detect weeds in various crops, including soybean \cite{soybean}, cereal crops \cite{dyrmann_cnn_cereal}, ryegrass \cite{yC_snn_ryegrass},  canola \cite{asad_cnn_canola} and rice \cite{barrero_cnn_rice}. These methods produced satisfactory results in distinguishing the weed from highly color contrasted soil background. However, with the turfgrass background, the weed detection problem becomes more challenging, and we are developing new methods here to improve detection performance.


With the increasing capability of detection networks such as Faster R-CNN~\cite{RCNN}, YOLO (You Only Look Once)~\cite{YOLO}, and SSD (Single Shot Detector)~\cite{SSD}, object detection against a highly-similar background can be achieved effectively. However, these object detection networks only provide bounding box output, which is not adequate for further field operation, especially localization. The localization problem can be partially addressed by segmentation networks such as Mask R-CNN~\cite{mask} and Deeplab~\cite{deeplab}, because these networks achieve finer image segmentation results for objects of interest. The problem with such methods is the tremendously high annotation cost, i.e. these networks often require pixel-wise precise ground truth for training, which is difficult and expensive for weed detection problems. 

\begin{figure}[htb!]
 \centering
 \includegraphics[width=2.5in]{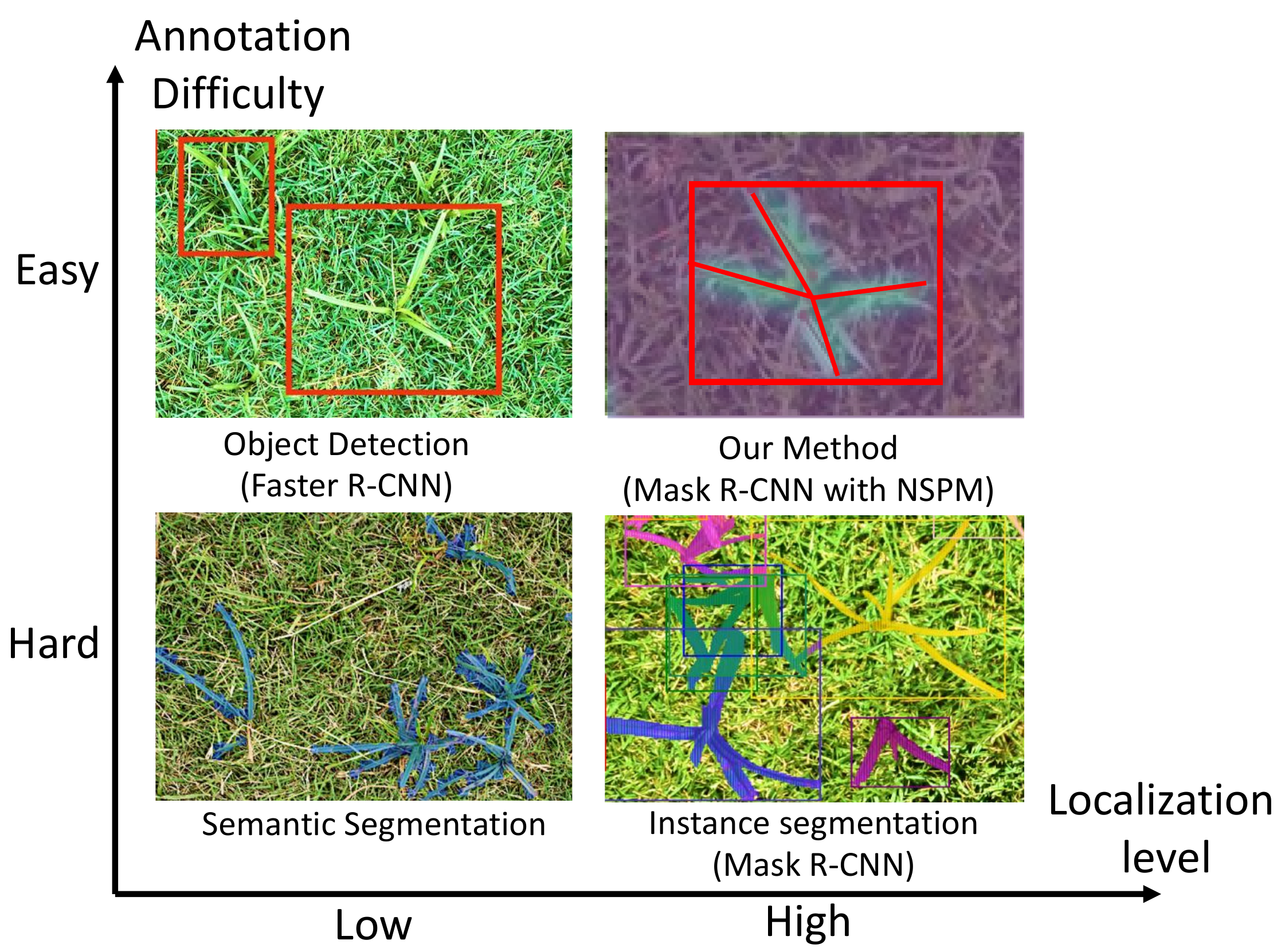}
\caption{Dimension of annotation difficulty and localization level for different methods.}
\label{fg:dim}
\end{figure}

Considering the unique shape of nutsedge leaves and plant architecture, extracting plant skeleton of nutsedge is a good approximation of semantic structure. In fact, the skeleton detection is also widely explored with end-to-end deep learning methods such as DeepFlux~\cite{DeepFlux} and Hi-Fi\cite{Hi-Fi}. Although these methods only target single object detection, which are not directly applicable in our scenarios, this inspires our development of nutsedge skeleton probability map to balance between the robustness of localization and annotation cost (Fig.~\ref{fg:dim}).   

\emph{Using synthetic data:} Researchers have explored different methods for data augmentation to enhance neural network training results, especially in domains where annotated data is difficult to obtain or expensive. Generative adversarial networks (GAN) is one of the methods that has gained popularity \cite{GAN_EYE}. However, training a GAN model to converge in specific tasks is often complicated and time consuming due to its adversarial nature. Thus, an easily accessible method for data augmentation is needed for nutsedge detection. Therefore, we synthesize images using real object segments. This approach involves a segmentation stage where nutsedge templates are extracted from real images either manually or automatically, and a synthesis stage where the extracted foreground nutsedge templates are pasted to the background of interest (i.e. turfgrass). Using a similar approach, Gao et al. \cite{gao_yolo_synthetic_images} train a YOLOv3 model for weed and crop detection, and achieve a mean average precision of 0.829. Toda et al.  \cite{toda_mask_rcnn_synthetic_images} show that a Mask R-CNN model for barley seed morphology phenotyping can be trained purely by a synthetically generated dataset where 96\% recall and 95\% average precision against real test dataset were achieved. Inspired by these results, we are developing a data-synthesis-based approach for weed detection problems.

\section{Problem Definition}\label{sc:Prob_def}

Our robot observes field through a downward facing camera to collect images (see video attachment for more details). Therefore, all images are collected from a perspective that is perpendicularly facing the ground from the same distance (0.5m in our set-up). 

Common notations are defined as follows:
\begin{itemize}
\item binary random variable $\mathbf{x}_{uv} = 1$ indicates event that pixel $(u, v)$ is a nutsedge pixel on the image where $u$ and $v$ are pixel indexes in horizontal and vertical directions, respectively.
\item $p(\mathbf{x}_{uv})$, probability of pixel at $(u,v)$ is a nutsedge pixel.  
\item $I_{r}:=\{(u,v): \forall (u, v) \}$, pixel set of a raw image collected from the field.
\item $P_o := \{ p(\mathbf{x}_{uv}): \forall (u,v) \in I_{r} ) \}$, a probability map set describing spatial probability distribution of $\mathbf{x}_{uv}$. It is the part of the output of the neural networks characterizing the confidence of the prediction.

\item $\mathbf{B}= \{B\}$ is a set of bounding boxes with each $B=\{(u,v)|u \in [u_{\mbox{\tiny left}}, u_{\mbox{\tiny right}}], v \in [v_{\mbox{\tiny bottom}}, v_{\mbox{\tiny top}}], (u,v) \in I_{r}\}$ where $(u_{\mbox{\tiny left}}, v_{\mbox{\tiny bottom}})  \in I_{r}$ and $(u_{\mbox{\tiny right}}, v_{\mbox{\tiny top}}) \in I_{r}$ is the bottom-left and top-right corners of the output bounding box, respectively. We use $\mathbf{B}_h$ representing human labeled bounding box set and $\mathbf{B}_o$ as algorithm output bounding box set.

\item $\mathbf{S}= \{S\}$ is a set of plant skeleton $S$ which will be defined later. We use $\mathbf{S}_h$ representing human labeled skeleton set and $\mathbf{S}_o$ as the algorithm output skeleton set.

\end{itemize}

The weed detection problem can be defined as follows,
\begin{Def}
Given the image collected by robot $I_{r}$, compute $\mathbf{B}_o$, $\mathbf{S}_o$ and $P_o$. 
\end{Def}

\section{Algorithms} \label{Algorithm}

Our algorithm development consists of three major components: data augmentation, network design \& training, and evaluation (Fig.~\ref{fg:pipeline}). Our data augmentation algorithm addresses the issue of insufficient training data by combining synthesised data with manually-labeled data. Due to the non-negligible level of errors existing in manually annotated labeling, we revised  the network design \& training to handle the inaccurately labeled training data. For the same reason, we cannot entirely trust the manually-labeled data as the ground truth and have to design a new evaluation pipeline considering the labeling noise to validate our model. We begin with the data augmentation.

\subsection{High Fidelity Data Augmentation}\label{ssc:data_augmentation}

\begin{figure}[htbp]
\centering
\includegraphics[width=3.5in]{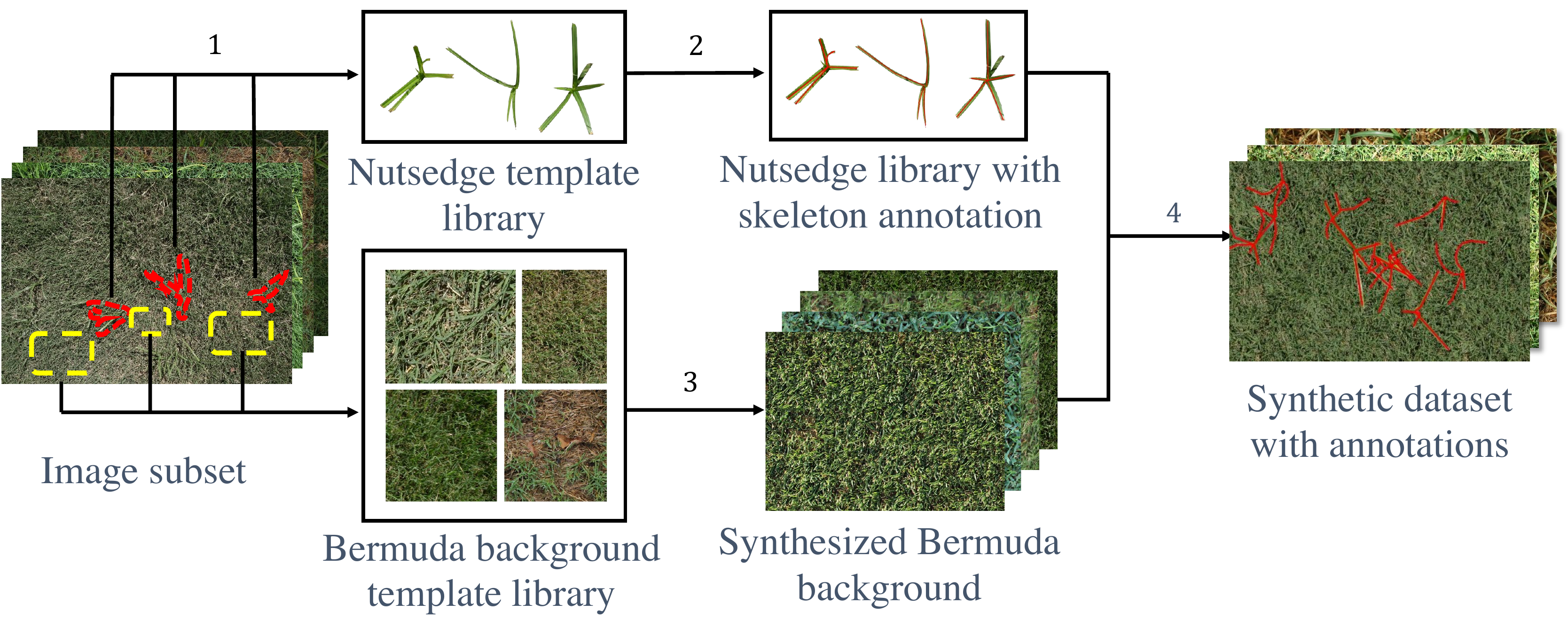}
\caption{An overview of the image synthesis pipeline. 1,2,3 and 4 represents template selection, annotation, background synthesis and recombination. Step 4 is the combined stage of image synthesis algorithm whose pseudocode is attached in the attached video file. 
}\label{fg:image_synthesis_pipeline}
\end{figure}

As detailed later, we employ deep neural networks for weed recognition which often require massive manually labeled data as the ground truth for training. We develop an image synthesis algorithm to efficiently generate high-fidelity artificial dataset from  images collected from the field with different granularity labels. Using data augmentation with image synthesis algorithm instead of directly labeling the raw images has three specific advantages: 1) it requires a minimal human labeling effort, 2) it expands the size of training dataset, and 3) it provides precise pixel-level labels.

In the synthetic dataset, each image is composed by nutsedge foreground and bermudagrass background. To generate realistic synthetic images with label, we ask human experts hand-select a small number of nutsedge templates and background patches from a raw image set as a material library (red and yellow dash shapes in the leftmost image in Fig.~\ref{fg:image_synthesis_pipeline}).  Then, image synthesis algorithm creates complete label sets for nutsedge template based on human expert's partial label for training purpose. The complete image synthesis algorithm consists of the following four steps corresponding to steps 1-4 in Fig.~\ref{fg:image_synthesis_pipeline}.   




\subsubsection{Template Selection}\label{ssc:templates} 
There are two libraries needed: a nutsedge template library (with skeleton and masking label) and a turf background library. To reduce the work load of human experts, we first employ the stratified random sampling \cite{stratified_sampling} based on the lighting condition to build an image subset (5\% of the training set) with  images under different lighting conditions proportional to raw image set. Human experts segment out nutsedge template $T \subset I_r$  where $T$ is a polygon covering nutsedge pixels, and nutsedge-free turf background pixel patches from the sampled image set.

\subsubsection{Nutsedge Annotation}  

For each nutsedge template $T$, there are 3 different types of annotation: plant skeleton $S$, binary mask $M_s$, and bounding box $B_T$.

To simplify the labeling process and better describe the structure attribute of nutsedges, we use plant skeleton labeling $S$ in our network design. As illustrated in Fig.~\ref{fg:probmap}(a), $S$ models each nutsedge plants as a cluster of line segments where each line segment depicts the center of a leaf,
\begin{equation}\label{eq:skeleton-model}
S := \{ \mathbf{l}_k: k=1,...,k_{\mbox{\tiny max}} \},
\end{equation}
where  $k_{\mbox{\tiny max}}$ is the total number of the line segments, and line segment $\mathbf{l}_k = \{ (u,v),(p,q)\}$, $(u, v)\in I_{r}$ and $(p, q)\in I_{r}$ are endpoints of the line segment. In the annotation process, one skeleton corresponds to one bounding box.  The line segments forming skeleton are annotated by human expert.  

The mask labels and  bounding boxes are generated automatically from nutsedge templates $T$. Following the manner of instance segmentation dataset creation \cite{coco}, $M_s$ labels are created by setting all template pixel as 1 for foreground nutsedge pixels and 0 otherwise. The bounding box computed from $T$ is defined as 

\begin{equation}\label{eq:boundbox_def}
B_{\tiny T} :=\{(u,v)|u \in [u_{\mbox{\tiny left}}, u_{\mbox{\tiny right}}], v \in [v_{\mbox{\tiny bottom}}, v_{\mbox{\tiny top}}], (u,v) \in I_{r}\},
\end{equation}
where $u_{\mbox{\tiny left}} = \min\{u\}$, $v_{\mbox{\tiny bottom}} = \min\{v\}$, $u_{\mbox{\tiny right}} =  \max\{u\}$, $v_{\mbox{\tiny top}}= \max\{v\}$, and $(u,v) \in T$.

\subsubsection{Background Synthesis}
To generate realistic background images with appropriate size and scale, a natural texture synthesis algorithm \cite{texture_synthesis} is employed. The advantage of using this algorithm over directly tiling with background templates is that it adds randomness to the synthesized background so as to prevent the neural network from picking up the unique patterns of each background template. 

\subsubsection{Recombination of Nutsedge and Background}
After background synthesis, the foreground of randomly selected subsets of the nutsedge template library are pasted onto the synthesized background images. The size of the subsets follows a uniform distribution within a desired range (this range is determined by experiment settings). While pasting each nutsedge template, the pixel locations in homogeneous coordinate are transformed by 2D coordinate transformation matrix 
$
\begin{bmatrix}
\cos(\theta) & \sin(\theta) &  t_{x}\\
-\sin(\theta) & \cos(\theta) &  t_{y} \\
0 & 0 & 1
\end{bmatrix}
$
where $\theta$ is a random rotation angle within $[0, 2\pi)$, and $t_{x}$ \& $t_{y}$ are horizontal and vertical random translations, respectively. They have uniformly distributed value within the image boundary. The resulting images are then augmented in hue, saturation, value (HSV) color space by randomly varying brightness value from 80$\%$ to 120$\%$ so that the trained models are more robust to the color variation in the testing dataset as a result of inconsistency for light conditions. The skeleton annotations of each nutsedge template are also inserted during the image synthesis process.

 The overall time complexity of the proposed image synthesis algorithm is $\mathcal{O}(u_{\mbox{\tiny max}} v_{\mbox{\tiny max}}s^2)$ where $(u_{\mbox{\tiny max}} v_{\mbox{\tiny max}})$ is the maximum image size in pixel count, and $s$ is the neighbourhood size for pixel candidate searching. In our implementation, the neighbourhood size $s$ is 24 pixels. The detailed pseudocode and anlysis is in the attached video file.


\subsection{Network Design and Training}
With both synthesized data and human-annotated training data (i.e. all raw image training set comes with human-labeled bounding boxes), we employ Mask R-CNN~\cite{mask} to develop our detector. In the original Mask R-CNN structure, the binary mask branch segments the image by assigning each pixel to a class. To better capture the feature of nutsedge while considering the imprecision in training dataset, we design a skeleton probability map representation of mask and modify the loss function of Mask R-CNN's mask branch correspondingly. 

\subsubsection{Nutsedge Skeleton-based Probabilistic Map Generation}

\begin{figure}[htbp!]   
\centering              
\includegraphics[width=0.5\textwidth]{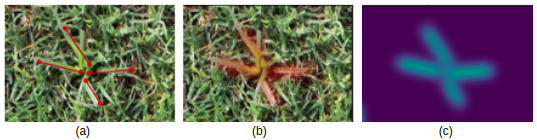} \caption{An example of NSPM: (a) skeleton from the data synthesis, (b) pixels masked as nutsedge in the synthesized image, and (c) the resulting nutsedge skeleton probability model.} 
\label{fg:probmap}
\end{figure}
For nutsedge segmentation problem, the difficulty of distinguishing the boundary of nutsedge's class increases as the distance from the center of nutsedge grows. Meanwhile, detecting the center and leaf midrib of the nutsedge is more important than detecting its edges for plant recognition. This motivates us to propose the use of NSPM input. The purpose is to instruct the network to differentiate the central leaf midrib part of the nutsedge, while reducing the impact of imprecision in nutsedge boundary segmentation.

Fig.~\ref{fg:probmap} illustrates NSPM computation. The bounding box for a skeleton $S$ is defined as
$B_h :=\{(u,v)|u \in [u_{\mbox{\tiny left}}, u_{\mbox{\tiny right}}], v \in [v_{\mbox{\tiny bottom}}, v_{\mbox{\tiny top}}], (u,v) \in I\}$ in similar format of $B_T$ in \eqref{eq:boundbox_def} with $u_{\mbox{\tiny left}}$, $v_{\mbox{\tiny bottom}}$, $u_{\mbox{\tiny right}}$, $v_{\mbox{\tiny top}}$ determined by human labeling instead of $T$.   For image $I$, we define the bounding box set as $\mathbf{B}_h = \{B_h\}$. 
For pixel $(u,v) \in B_h$ which contains the plant skeleton $S$, the probability of $(u,v)$'s class is nutsedge
\begin{equation}\label{eq:ps}
\nonumber p_S(\mathbf{x}_{uv} )  \propto  
\begin{cases}
\sum_{k=1}^{k_{\mbox{\tiny max}}} & \frac{1}{\sigma \sqrt{2\pi}}\exp\{-\frac{1}{2} [\frac{d((u,v), \mathbf{l}_k)}{\sigma }]^2\}  \\
 ~~~&\mbox{if }((u,v) \in B_h)\land (B_h\in \mathbf{B}_{h}) \\
0,& \mbox{~otherwise.} 
\end{cases}
\end{equation}
where $d((u,v),\mathbf{l}_k)$ is the point $(u,v)$ to line segment $\mathbf{l}_k$'s nearest point's distance, and we use the nutsedge template  and the skeleton labeled by human expert to estimate the proper value of $\sigma$. By drawing the histogram for each nutsedge template  in the template library with $d((u,v),\mathbf{l}_k)$ as x-axis and count of pixel number as y-axis, we can use half-normal distribution to approximate the histogram and estimate $\sigma$.

The NSPM $P_S$ of the image is defined as
\begin{equation}\label{eq:PS}
P_S(u,v) = \{p_S(\mathbf{x}_{uv}), \forall (u,v) \in I \}
\end{equation}
$P_S$ is used as the annotation input for the training image $I$. 

\subsubsection{Modifying Loss Function}
At the same time, we need to modify the original loss function (cross entropy) in mask branch to accommodate labeling imprecision. In original loss function, it maps origin binary annotation (ground-truth) value to discrete distribution for binary mask $M_s$ as $p_1(\mathbf{x}_{uv})\in \{ 0,1 \}$ and represents mask branch output as probability density function $p_2(\mathbf{x}_{uv})\in [0,1]$

\begin{equation}\label{eq:cross_entropy}
 L_{H}(p_1,p_2) =-\sum_{(u,v)\in B_o} p_1 (\mathbf{x}_{uv}) \log(p_2 (\mathbf{x}_{uv})).
\end{equation}

The problem of cross entropy loss function is that it is designed for deterministic annotation without considering the uncertainty introduced by the imprecision in labeling. To address this problem, we introduce KL-Divergence as the loss function for mask branch that perceives the uncertainty in human annotation and model it as a probability distribution using NSPM, where the annotation's probability distribution is $p_1(\mathbf{x}_{uv})= P_S(u,v)$. 

 \begin{equation}\label{eq:D_KL}
 L_{KL}(p_1,p_2) =-\sum_{(u,v)\in B_o} p_1 (\mathbf{x}_{uv}) \log(\frac{p_1 (\mathbf{x}_{uv})}{p_2 (\mathbf{x}_{uv})}).
 \end{equation}

 It is worth noting that \eqref{eq:cross_entropy} and \eqref{eq:D_KL} share the same time complexity in computation. When we replace the cost function \eqref{eq:cross_entropy}  with \eqref{eq:D_KL}, our revised Mask R-CNN share the same time complexity with the original version.

\subsubsection{Transfer Learning Using Data with Different Levels of Annotation} A worth-mentioning design of our training dataset is that images have labels at different levels of granularity. Human labeled raw image set only contains the bounding box annotation, while the synthesized data generated by nutsedge template have higher precision level labels: binary mask and plant skeleton label.  

To efficiently train our model with different annotation levels, we develop a new training strategy for Mask R-CNN. As an instance segmentation network, Mask R-CNN outputs the bounding box, the class of bounding box, and the binary mask of nutsedge. All the three branches share the same backbone feature extraction and Region Proposed Network (RPN) \cite{RCNN}. Our training strategy fully exploits the structure's potential. First, we employ raw image $I_r$ with human labeled bounding box $B_h$ to train the model's classification and bounding box detection branch to ensure that the feature extraction network has been mostly trained from real data's distribution and human observation (Fig. \ref{fg:pipeline}, dash line's flow). Second, we fine-tune the feature extractor and train the original mask branch using synthesized data $I_s$ with its label $M_s$ (Fig. \ref{fg:pipeline}, 
solid line's flow).

\subsubsection{Skeleton Decoder} 
When we train the Mask R-CNN, we adopt ResNet-FPN\cite{FPN} backbone to obtain feature fusing map in the feature extraction stage. With the high-resolution and high-level semantic map embedded in the same feature map, the model learns complex semantic information through training. The inference output probability map $P_o$ has a higher probability in the midrib of leaves. This attribute of the probability map enables us to extract nutsedge skeleton from it. After receiving the probability map $P_o$, we adopt pre-processing morphology dilation and erosion with the Gaussian blur to make the probability map distribution more smooth. Then, we apply a non-maximum suppression skeleton selection \cite{Hi-Fi} algorithm to the pre-processed probability map to decode its skeleton structure.

\subsection{Semi-supervised Evaluation}
\label{eval}

Standard evaluation methods for detection and segmentation problem often compare the region similarity using intersection over union (IoU) metric between the model output and label of the bounding box (ground-truth). As we described early, due to the labeling imprecision, human annotation cannot be treated as ground truth. Thus, a new evaluation method is needed. Here we design evaluation methods targeting situations when human annotations and model are consistent or inconsistent, respectively.

\subsubsection{Consistent Metrics}
\label{SM}
In this step, we evaluate how model outputs compare to bounding box set labeled by human ($\mathbf{B}_h$) when they are consistent. For this purpose, we compare both pixel-wise region overlap and skeleton similarity.

\begin{itemize}
\item\textbf{Region overlap:} 
With human labeled bound box $\mathbf{B}_h$ set and skeleton $S_h$ set, we can obtain probability map $P_S$ using \eqref{eq:PS}. We can threshold $P_S$ to obtain region set $I_S$ according to human labels,
\begin{equation}\label{eq:I_s_def}
    I_S:=\{(u,v)| p_S(\mathbf{x}_{uv})>t\} \subseteq I_r,
\end{equation}
where $t$ is probability threshold. Similarly, we can obtain region set $I
_o$ according to the model output probability map $P_o$ using the same threshold. The region overlap between $I_S$ and $I_o$ can be measured by IoU metric,
\begin{equation}\label{eq:IoU}
r_{\mbox{\tiny IoU}} = \frac{|I_S \cap I_o|}{|I_S\cup I_o|},
\end{equation}
where $|\cdot|$ is set cardinality.

\item \textbf{Skeleton similarity:} We use the skeleton similarity between $S_o$ and $S_h$ to evaluate how well the model capture main structure of the nutsedge. First, for each pixel $(u, v)$ in $S_o$, if we can find the distance $d_{S_h}(u, v)$ to its closest point in $S_h$,
\begin{equation}
    d_{S_h}(u, v) = \min_{(u_a, u_b) \in S_h}\sqrt{(u_a-u)^2+(u_b-v)^2)}.
\end{equation}
If $d_{S_h}(u, v)$ is less than a given threshold $d$, we believe that the pixel $(u, v)$ has a corresponding point in $S_h$. We obtain the ratio between the corresponding pixel counts in $S_h$ and the total pixel number in $S_h$,
\begin{equation}\label{eq:skeleton_similarity}
    C_s = \frac{ |\{(u, v)| (u, v) \in S_o, 
     d_{S_h}(u, v) \leq d \} |}{|S_h|}  
\end{equation}
as the skeleton similarity metric.
\end{itemize}

\subsubsection{Inconsistent Metrics} 

For our problem, it is possible that the model fails to recognize a nutsedge and it is also possible that human may make mistakes in annotation. We want to catch these inconsistent cases and further analyze them. 

First, we identify the consistent bounding box set  $\mathbf{R}_a$,
$$\mathbf{R}_a = \{ B \mid B\in \mathbf{B}_h \cap \mathbf{B}_o, (r_{\mbox{\tiny IoU}} \ge 0.5) \lor (C_s \ge 0.7) \},$$
where $r_{\mbox{\tiny IoU}}$ and $C_s$ are computed using \eqref{eq:IoU} and \eqref{eq:skeleton_similarity} respectively. Then we obtain the inconsistent bounding box set $\mathbf{R}_c = \{ (\mathbf{B}_h \cup \mathbf{B}_o) \setminus \mathbf{R}_a \}$. 
When inconsistent cases are detected, we manually reexamine the labels of those bounding boxes and classify $\mathbf{R}_c$ into three groups: 1) false positive case set of algorithm output $\mathbf{B}^o_{\mbox{\tiny FP}}$, 2) false negative case set of algorithm output $\mathbf{B}^o_{\mbox{\tiny FN}}$, and false negative set of human annotation $\mathbf{B}^h_{\mbox{\tiny FN}}$.

\section{Experiment}

We have implemented our weed detection algorithm based on Detectron2\cite{wu2019detectron2} system on Pytorch platform. We choose ResNet-50 with Feature Pyramid Networks (FPN) and ResNet-101 with FPN as our backbone network. The initial network parameters of Faster R-CNN and Mask R-CNN are both from a pre-trained model on MSCOCO dataset \cite{coco}. 

\subsection{Nutsedge dataset}
We have built the a shared TAMU nutsedge dataset \cite{weed-data2021} which contains two types of data: the raw image set collected from the field with manual annotations and synthetic image set with ground truth synthetic label.   

\subsubsection{Raw Image Set}
The raw images were collected at the Scotts Miracle-Gro Facility for Lawn and Garden Research, Texas A\&M University using Nikon\texttrademark~D3300 or Canon EOS Rebel T7\texttrademark~mounted at a height of 0.5m on a data collection cart. See attached video file for more details. The original image resolution is $6000\times4000$ but  downsized to $1200\times800$ to adapt the model and reduce training costs. To cover the appearance variation of nutsedge, data are collected at different lighting conditions, temperature, weather, and moisture levels. To cover the majority of nutsedge growth season, data were collected from June to August at different times of day. The raw dataset contains 6000 images which is split into a training set $D_r$ (90\%) and a testing set $D_{t}$ (10\%). All data are labeled with bounding boxes for both training and testing purposes. In addition, 25\% of the testing images contain skeleton label. We denote the testing set with skeleton label as $D_{t_S} \subseteq D_{t} $. The size of $D_{t_S}$ is $n_{t_S}=|D_{t_S}|$. All the labels are created by human annotation using ``labelme"\cite{labelme2016} tool.

\subsubsection{Synthetic Dataset} Generated using the method in Section~\ref{ssc:data_augmentation}, our synthetic dataset contains 4750 images with bounding box labels, which are used as the training set. The density of nutsedges is set at 5 to 10 plants per one million pixels. When we generated the NSPM, we set $\sigma = 12$ pixel based on statistical analysis of existing data. Moreover, the dataset contains both binary mask label and skeleton label. When only the binary pixel-level mask label is used with the synthetic dataset, we name it as $D_{s_b}$. When only skeleton label is used with the synthetic dataset, we name it as $D_{s_p}$. $|D_{s_b}| = |D_{s_p}| = 4750$.
The sample images of synthesized dataset is shown in attached multimedia file. 

\subsubsection{Reduction of Labeling Time} 
The data synthesis algorithm significantly reduces manual labeling effort. The average density of nutsedge in raw dataset is 10 plants per image. It takes about 30 seconds to label for each plant. To label all 800 raw images with mask label, it would cost 66 hours. With the help of image synthesis algorithm, we only need to select and create mask label for 129 nutsedge template. The labeling time is reduced to less than 3 hours which is a 95\% reduction in labeling time. Also, the synthetic data contains ground truth that is not attainable in noisy manually labeled data.


\subsection{Component Tests}

\subsubsection{Loss Function Comparison}
\begin{figure}[htb!]
 \centering
 \includegraphics[width=3in]{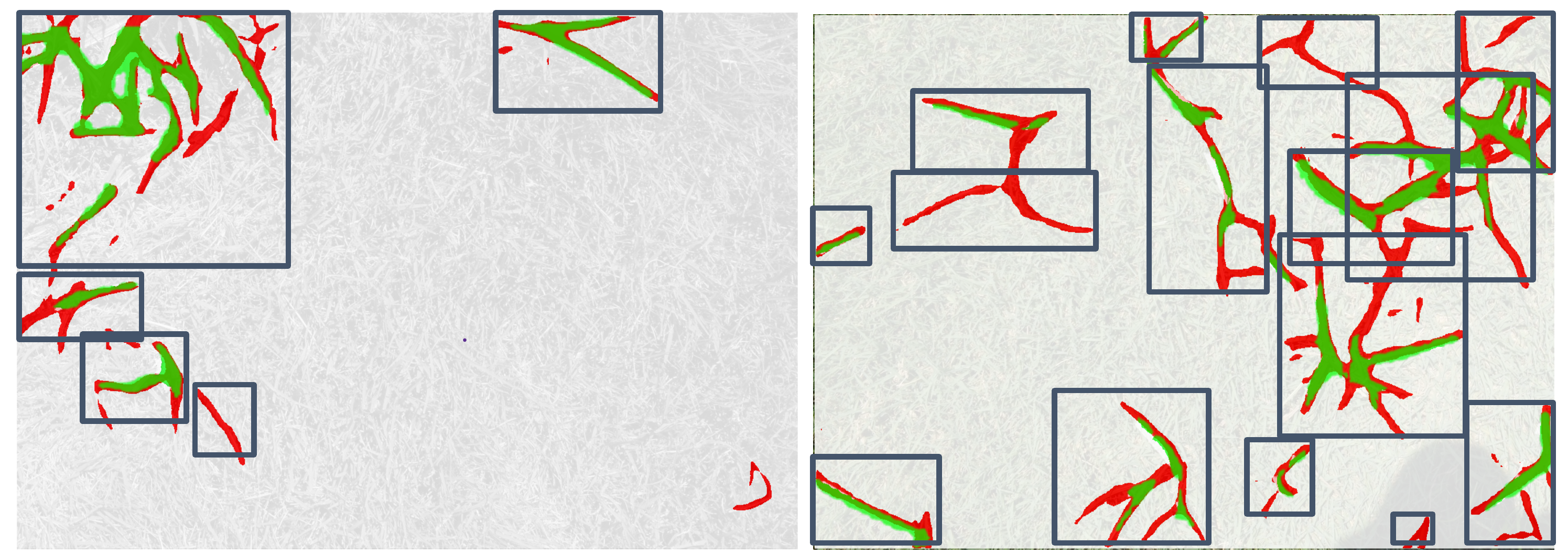}
\caption{A comparison of the detection results with cross entropy (in green) and KL-divergence (in red) models. The grey boxes are bounding boxes from manual labeling. It is clear that there are a lot more red pixels than green pixels, which means that the KL-divergence loss function misses fewer than the cross entropy loss function. Both models use R101 as the backbone.
}
\label{fg:exp_fig}
\end{figure}

We train a Mask R-CNN model using cross entropy loss function with dataset $D_{s_b}$ and using KL-divergence with dataset $D_{s_p}$. 
The $r_{\mbox{\tiny S-IoU}}$ is an average $B_h$'s $r_{\mbox{\tiny IoU}}$ in an image weighted by skeleton size. We calculate the $\overline{r}_{\mbox{\tiny IoU}}$ by averaging all image's $r_{\mbox{\tiny S-IoU}}$. Let $n_b = |\mathbf{B}_h|$ in one image and the total pixel count of skeleton in the image be $c_{I_S}=\sum_{n_b} |S_h|$. We have
\begin{equation}\label{eq:IoUs}
    r_{\mbox{\tiny S-IoU}} = \frac{1}{n_b}\sum_{n_b} \frac{|S_h|}{c_{I_S}}r_{\mbox{\tiny IoU}} ~~\mbox{and}~~
     \overline{r}_{\mbox{\tiny IoU}} = \frac{1}{n_{t_S}}\sum_{n_{t_S}} r_{\mbox{\tiny S-IoU}}.    
\end{equation}
Similarly, we extend the skeleton similarity metric, 
\begin{equation}\label{eq:skeleton_sim_averags}
C_{Ss} =  \frac{1}{n_b}\sum_{n_b} \frac{|S_h|}{c_{I_S}} C_s ~~\mbox{and}~~
\overline{C}_s = \frac{1}{n_{t_S}}\sum_{n_{t_S} }C_{Ss}.
\end{equation}
The overall result is shown in Table.~\ref{loss_table}. We use R50, R101, CE and KL representing the ResNet-50-FPN, ResNet-101-FPN, cross entropy and KL divergence, respectively. It is clear that changing the loss function from CE to KL achieves higher $\overline{r}_{\mbox{\tiny IoU}}$ and $\overline{C}_s$. Even with a smaller backbone network (R50), the model trained by KL loss function performs better than that by R101 using CE loss function by 3\% in $\overline{r}_{\mbox{\tiny IoU}}$ and 4\% in $\overline{C}_s$. When the backbone is identical, the model with KL loss improves over the CE by more than 10\% in both $\overline{r}_{\mbox{\tiny IoU}}$ and $\overline{C}_s$. Sample results are shown in Fig.~\ref{fg:exp_fig}.  

\subsubsection{Improvement with Transfer Learning} \label{tranfer_exp}
We follow the basic rules of transfer learning by using the pre-trained model to improve the performance. In general case, without the task-specialized pre-trained model, the common models such as the one trained by MSCOCO is used as the pre-trained model. The first four lines in Table~\ref{loss_table} use MSCOCO pre-trained model as initial parameters. To get further improvement, we use $D_r$ to pre-train the backbone and bounding box branch. The performance of model with $D_r$ pre-trained and R101 as backbone lists in the line 5 of Table~\ref{loss_table} which is highlighted in bold font as the best performer.

\begin{table}[htbp!]
    \centering
    \caption{Detection comparison.}
    \setlength{\tabcolsep}{3.2mm}
    \begin{tabular}{cccccc}
    \hline
    Training& Testing & Backbone & Loss & $\overline{r}_{\mbox{\tiny IoU}}$ & $\overline{C}_s$\\
    \hline
    $D_{s_b}$ & $D_{t_S}$ & R50 & CE & 0.42 & 0.75\\
    $D_{s_b}$ & $D_{t_S}$ & R101& CE & 0.45 & 0.77\\
    $D_{s_p}$ & $D_{t_S}$ & R50& KL & 0.48 & 0.81\\
    $D_{s_p}$ & $D_{t_S}$ & R101&KL & 0.57 & 0.88\\
     $D_{s_p}\cup D_{r}$ & $D_{t_S}$ & R101& KL& \textbf{0.61} & \textbf{0.88}\\
    \hline
    \end{tabular}
    
    \label{loss_table}
\end{table}

\begin{figure}[htbp]   
\centering          \includegraphics[width=3.5in]{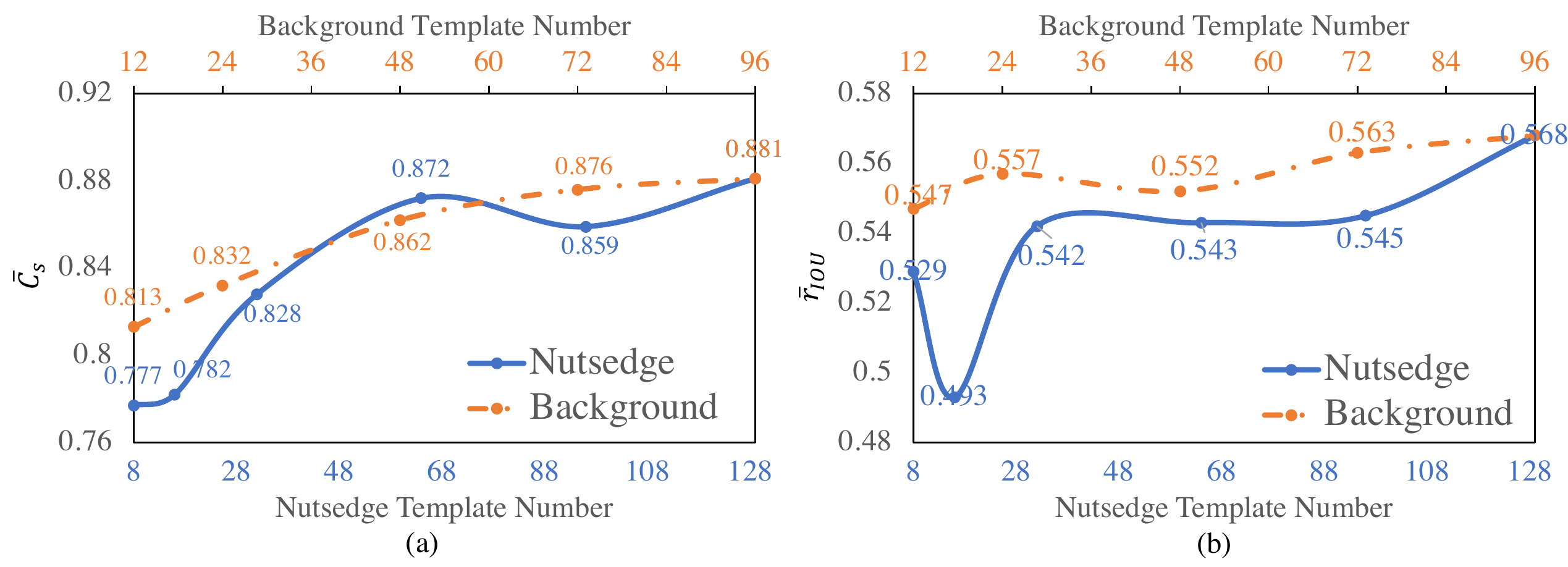}
\caption{Affect of different number of nutsedge and background templates in generating synthetic training data.}\label{fg:template_experiment} 
\end{figure}
\subsubsection{Synthetic Data Generation Configuration} Synthetic data provides accurate ground truth with pixel-level mask, which is expected to substantially improve the model. We study how the number of foreground nutsedge and background bermudagrass templates (Section~\ref{ssc:templates}) in generating synthetic data affects the overall detection performance. 
First, we vary nutsedge foreground template sizes while keeping the background template number at 96. We increase the number of nutsedge templates from 8 to 129. Again, $\overline{r}_{\mbox{\tiny IoU}}$ and $\overline{C}_s$ are used to evaluate the detection result (Fig.~\ref{fg:template_experiment}). With mere 8 nutsedge templates, the trained model achieves $\overline{r}_{\mbox{\tiny IoU}}$ of 52.9\% and $\overline{C}_s$ of 77.7\%. As the nutsedge templates increase, the $\overline{r}_{\mbox{\tiny IoU}}$ gradually grows to 56.8\% and $\overline{C}_s$ reaches 88.1\%. Similarly, we test our algorithm by changing the background template number from 12 to 96 while fixing the number of nutsedge templates at 129. $\overline{r}_{\mbox{\tiny IoU}}$ and $\overline{C}_s$ are 54.7\% and 81.3\%. The curve in Fig.~\ref{fg:template_experiment} also illustrates the positive correlation between the number of background templates and the model performance, but the trend is relatively less significant compared to that of the nutsedge template number. Considering the fact that selecting templates is expensive, we choose 129 nutsedge templates with 96 background templates as our setup in generating synthetic data.

\subsection{Overall Performance Comparison}

\begin{table}[htbp]
    \centering 
    \caption{Overall performance comparison}
    \begin{tabular}{cccccccc}
    \hline
    Alg. & Loss & Training set & $r_d$  & $r_a$ & $r_{\mbox{\tiny FN}}$ & $r_{\mbox{\tiny FP}}$ \\
    \hline
     $a$ & CE & $D_r$ & 3.01 & - & - &  -  \\
    $b$ & CE & $D_{s_b}$ & \textbf{22.71} & 94.3\% & 5.0\% & \textbf{0.2\%} \\
    $c$ & KL & $D_{s_p}$ & 21.14 & 96.8\% & 0.7\% & 1.7\%  \\
    $d$ & KL  & $D_{s_p}\cup D_r$ &18.91& \textbf{97.1\%} & \textbf{0.4\%} & 4.4\% \\

    \hline
    \end{tabular}

    \label{tb:overall_exp}
\end{table}

\subsubsection{Algorithms and Training Setup}
The overall evaluation compares the four algorithms indicated below (Algs. a-d)  under their required training setup. In fact, Algs. c-d are our algorithms with different configuration.
\begin{itemize}
    \item[a.] Faster R-CNN based model with R101 backbone: this setup only uses bounding boxes as training set input and algorithm output, and it does not require pixel-level labeling  A sample input is shown in the top left image in Fig.~\ref{fg:dim}. This algorithm serves as a baseline for Faster R-CNN. 
    \item[b.] Mask R-CNN based model with R101 as the backbone and trained by CE loss function: Here we use synthetic data with binary pixel-level mask label $D_{s_b}$.  This algorithm tests the power of synthetic data and can also be viewed as an approximate baseline for Mask R-CNN with precise labeling.  A sample input is shown in bottom right image in Fig.~\ref{fg:dim}. The typical application of the original Mask R-CNN would require fully manually labeled pixel-wise training data. In fact, the synthetic data remove labeling noise which may make the algorithm performs better than the actual case. Also, the synthetic data may not be as representative as the precise real data which are not available. That is reason for us to call it an approximate baseline for Mask R-CNN. The comparison is not exact but still meaningful.
    \item[c.] We change Alg. b settings by swapping the loss function from CE to KL divergence in~\eqref{eq:D_KL}. The swapping also allows us to use skeleton-labeled synthetic set $D_{s_p}$. This algorithm examines if the change of loss function improves the performance.
    \item [d.] We further extend the model c with a pre-trained model described in Sec.~\ref{tranfer_exp}. Also, real training set $D_r$ is used in combination with $D_{s_p}$. This algorithm is presumed to be the best overall according to the component test. 
\end{itemize}
All models are tested on the raw image set $D_{t_S}$.  

\subsubsection{Metrics and Results} \label{overall_performance}
To compare the detection ability of algorithm with only bounding box output (a) and Algs. with precise pixel-level output (b-d), we define the density ratio $r_d$ as the ratio between nutsedge density of detection region and density of the entire image:
$$r_d = \frac{c_a/c_o}{c_s/c_I},$$
where  $c_s$ is the total number of nutsedge pixels, $c_I$ is the total pixel count of the testing image, $c_a$ is the total number of nutsedge pixels covered by output bounding boxes, and $c_o$ is the pixel count for the union area of the output bounding boxes. $c_s$ and $c_a$ are based on human labeling results since Alg. a's input and output are just bounding boxes. High values of $r_d$ indicate better detection because the algorithm is able to identify focused regions with more nutsedges. Table~\ref{tb:overall_exp} shows the result. It is clear that Algs. b-c perform much better than Alg. a. This is expected because raw image with human label contains high error in training samples, which negatively affect detection results. For Algs. b-c, the use of synthetic data greatly improves network training. 

For Algs. b-c, $r_d$ does not tell the complete story. We need to take a closer look because not all nutsedge pixels are equal or error-free. Further, we are also interested if disagreements between algorithm and human labeling can reveal more insights. To focus on this, we need new metrics that do not simply treat human label as ground truth. Let $\mathbf{N}_d$ be the total detected nutsedge bounding box set based on both algorithm output and human labeling. It is a union of consistent case $\mathbf{R}_a$, cases missed by model output $\mathbf{B}^o_{\mbox{\tiny FN}}$, and cases missed by human label $\mathbf{B}^h_{\mbox{\tiny FN}}$:
$\mathbf{N}_d = \{ \mathbf{R}_a \cup \mathbf{B}^o_{\mbox{\tiny FN}} \cup \mathbf{B}^h_{\mbox{\tiny FN}} \}$. It is worth noting that these metrics build on segmented nutsedge pixels (i.e. region overlap in \eqref{eq:IoU} and skeleton similarity \eqref{eq:skeleton_similarity}). Cases outside $\mathbf{R}_a$ are subjected to a manual re-examination step to determine  which ones are correct. These metrics do not apply to Alg. a due to its lack of segmentation capability. For the rest, these sets allow us to define the agreement rate $r_a$, false positive rate of model $r_{\mbox{\tiny FP}}$ and false negative rate of model $r_{\mbox{\tiny FN}}$ as model comparison metrics. 
$$r_a = \frac{|\mathbf{R}_a|}{|\mathbf{N}_d|},~~  r_{\mbox{\tiny FP}} = \frac{|\mathbf{B}^o_{\mbox{\tiny FP}}|}{|\mathbf{N}_d \cup \mathbf{B}^o_{\mbox{\tiny FP}}|}, ~\mbox{and}~r_{\mbox{\tiny FN}} = \frac{|\mathbf{B}^o_{\mbox{\tiny FN}}|}{|\mathbf{N}_d|}. 
$$
Table~\ref{tb:overall_exp} shows that Alg. d achieves the best overall results. This is due to high overall agreement between human and algorithm output and the lowest false negative ratio. Algorithms with low false negative detection help remove weeds more thoroughly. However, in situations where herbicide use reduction is much more important than thorough weed control, we may want to choose Alg. b due to its lowest false positive rate.

\section{Conclusion and Future work}

We reported our weed detection algorithm development for robotic weed control. We focused on detecting nutsedge weed in bermudagrass turf. Building on the Mask R-CNN, an instance segmentation framework, our new algorithm incorporated four new designs to handle the imprecision and insufficiency of training datasets. First, we proposed a data synthesis method to generate high fidelity synthetic data. We combined the precise labeling from the synthetic data and noisy labeling from the raw data to train our network. We also proposed new data representation to allow the network to focus on the skeleton of the nutsedge instead of individual pixels. We modified the loss function to enable Mask R-CNN to handle training data with high uncertainty. We also proposed new evaluation metrics to facilitate comparison under imprecise ground truth. The experimental results showed that our design was successful and significantly better than the Faster R-CNN approach. 

In the future, we will extend our approach to more types of weeds and turf species. Building on these results, we will also develop robotic weed removal algorithms and systems, and test them under field conditions.

\section*{Acknowledgment}
\footnotesize{The authors would like to thank H. Cheng, S. Yeh, A. Kingery, A. Angert, and D. Wang for their early feedback and  contributions to the NetBot Lab at Texas A\&M University. We acknowledge Z. Mansour and B. Wherley for assisting with field image collection. This research was funded by a T3 grant provided by Texas A\&M University.}

\bibliographystyle{IEEEtran}
\bibliography{citation}

\end{document}